%% file: acl_latex.tex
\pdfoutput=1

\documentclass[11pt]{article}

\usepackage{acl}

\usepackage{times}
\usepackage{latexsym}
\usepackage{xspace}

\usepackage[T1]{fontenc}

\usepackage[utf8]{inputenc}

\usepackage{microtype}

\usepackage{inconsolata}
\usepackage{subfig}
\usepackage{hyperref}
\usepackage{url}
\usepackage{caption}
\usepackage{pdfpages}
\usepackage{tikz}
\usepackage{algorithm}
\usepackage{pifont}
\usepackage{multirow}
\usepackage{algorithmicx}
\usepackage{algpseudocode}
\usepackage{booktabs}
\usepackage{pgfplots}
\usepackage{graphicx}
\usepackage{booktabs}
\usepackage{color}
\usepackage{mathtools}
\usepackage{dsfont}
\usepackage{array}
\usepackage{xcolor,colortbl}

\usepackage{listings}

\definecolor{dkgreen}{rgb}{0,0.6,0}
\definecolor{gray}{rgb}{0.5,0.5,0.5}
\definecolor{mauve}{rgb}{0.58,0,0.82}

\input{math_commands.tex}

\newcolumntype{C}[1]{>{\centering\arraybackslash}p{#1}}

\definecolor{Gray}{gray}{0.85}
\definecolor{LightCyan}{rgb}{0.88,1,1}
\newcolumntype{g}{>{\columncolor{Gray}}c}
\newcolumntype{w}{>{\columncolor{white}}c}

\title{ToolNet: Connecting Large Language Models \\ with Massive Tools via Tool Graph}

\author{Xukun Liu \\
  Northwestern University \\
  \hspace{-15mm}
  \texttt{xukunliu2025@u.northwestern.edu} \\\And
  Zhiyuan Peng \\
  North Carolina State University \\
  \texttt{jerrypeng1937@gmail.com} \\\And
  Xiaoyuan Yi \\
  Microsoft Research Asia\\
  \texttt{xiaoyuanyi@microsoft.com} \\\AND
  Xing Xie\\
  Microsoft Research Asia\\
  \texttt{xingx@microsoft.com} \\\And
  Lirong Xiang\\
  North Carolina State \\University \\
  \texttt{lxiang3@ncsu.edu} \\\And
  Yuchen Liu \\
  North Carolina State\\ University \\
  \texttt{yuchen.liu@ncsu.edu} \\\And
  Dongkuan Xu\\
  North Carolina State\\ University \\
  \texttt{dxu27@ncsu.edu} \\
  }

\begin{document}
\maketitle
\begin{abstract}
While achieving remarkable progress in a broad range of tasks, large language models (LLMs) remain significantly limited in properly using massive external tools. Existing in-context learning approaches simply format tools into a list of plain text descriptions and input them to LLMs, from which, LLMs generate a sequence of tool calls to solve problems step by step. Such a paradigm ignores the intrinsic dependency between tools and offloads all reasoning loads to LLMs, making them restricted to a limited number of specifically designed tools. It thus remains challenging for LLMs to operate on a library of massive tools, casting a great limitation when confronted with real-world scenarios. This paper proposes \sysname, a plug-and-play framework that scales up the number of tools to thousands with a moderate increase in token consumption. \sysname organizes tools into a directed graph. Each node represents a tool, and weighted edges denote tool transition. Starting from an initial tool node, an LLM navigates in the graph by iteratively choosing the next one from its successors until the task is resolved. Extensive experiments show that \sysname can achieve impressive results in challenging multi-hop tool learning datasets and is resilient to tool failures.
\end{abstract}

\section{Introduction}
\input{sections/intro}

\section{Problem Formulation}
\input{sections/background}

\section{ToolNet}
\input{sections/method}

\section{Experiments}
\input{sections/exp}
\section{Related Work}
\input{sections/related}

\section{Conclusion}
\input{sections/conclusion}

\section{Limitations and Future Work}
\input{sections/limitation}





\bibliography{acl_latex}

\newpage
\section*{\Large Appendix}
\input{sections/appendix}



\end{document}

%% file: math_commands.tex
\usepackage{amsmath,amsfonts,bm}

\def\eqref#1{equation~\ref{#1}}

\def\1{\bm{1}}

\DeclareMathAlphabet{\mathsfit}{\encodingdefault}{\sfdefault}{m}{sl}
\SetMathAlphabet{\mathsfit}{bold}{\encodingdefault}{\sfdefault}{bx}{n}

\newcommand{\sysname}{\textit{ToolNet}\xspace}

%% file: sections/intro.tex
\begin{figure}[h]
\centering
\includegraphics[width=0.45\textwidth]{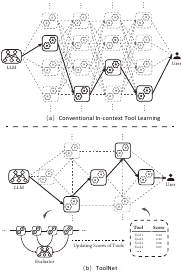}
\caption{
Comparison of (a) prior in-context tool learning methods and (b) the proposed \sysname. Conventionally, all tools are formatted as the input to an LLM for step-by-step tool calling, posing scalability challenges for using massive tools. \sysname organizes tools into a graph. An LLM chooses only from the successor tools relative to its previous selection. An Evaluator assesses the effectiveness of tools and dynamically modifies the tool transition weights within the graph.
}
\label{fig:react_agentnet_overall}
\end{figure}
There is an emerging interest~\cite{qin2023toolllm,song2023restgpt,yao2022react,patil2023gorilla,yang2023mm} in unleashing the power of large language models (LLMs) to effectively interact with various tools (or APIs) for real-world tasks. Tool-augmented LLMs, when connected feasibly with massive APIs, can serve as an intermediate interface between average humans and complex tools, which may eventually reshape the vast ecosystem of applications. This endeavor has yielded several impressive industrial outcomes, such as New Bing the search engine~\cite{NewBing2023}, Copilot the office assistant~\cite{MicrosoftCopilot2023}, RT-2 the robot controller~\cite{brohan2023rt}, and WebGPT the web-browsing agent~\cite{nakano2112webgpt}. 



Despite the remarkable progress, tool-augmented LLMs are still in the experimental stage and not yet ready to fully meet real-world demands. Notably, while LLMs are designed as generalists for multiple tasks, LLM-powered agents are commonly customized with few-shot in-context examples for narrow purposes. They are limited to connecting with a small number of specially designed tools. For example, Toolformer~\cite{schick2023toolformer} masters 5 tools such as calculator, question-answering engine, calendar, etc. Chameleon~\cite{lu2023chameleon} specializes in two knowledge-intensive question-answering tasks with a set of well-curated tools, such as table verbalizer and image captioner.

Scaling up the number of tools to thousands is challenging for LLMs. This is not merely due to the surging token consumption when tools are prompted as input to LLMs, which commonly exceeds their token limits. More importantly, it is beyond LLMs' capability to select the correct one(s) from a library of vast tools through straightforward in-context learning. According to ~\cite{hao2023toolkengpt}, as the number of tools increases, LLMs tend to hallucinate and make mistakes in calling tools, causing steady performance degradation. In response, researchers put huge efforts into training LLMs to master massive tools~\cite{qin2023toolllm,patil2023gorilla,hao2023toolkengpt}. While this method could yield promising results, it brings prohibitive computation costs and lacks adaptability to new tools or tools with constant function updates. 
Moreover, in a vast tool library, low-quality tools should indeed be reckoned with. When misled by tools, LLMs are prone to hallucinate. In response, ~\cite{xie2023decomposition} employs beam search to explore the most proper tool(s). However, token consumption is multiplied, and without an explicit memory mechanism, the exploration experience cannot be leveraged for subsequent tasks.
To address the issues above, we raise the following questions:

\textit{\textbf{Question 1.} How to enable an LLM to cope with massive tools while retaining token efficiency?}

We analyze the multi-hop tool-use trajectories in ToolBench~\cite{qin2023toolllm}, 
recognized as the most extensive publicly available dataset dedicated to tool learning. 
As depicted in Figure~\ref{fig:toolbench_data}, our analysis reveals that tools generally have a restricted set of potential successor tools to be invoked. This observation signifies the sparse transition between tools. In essence, when a particular tool is invoked, the subsequent tool to be called can be constrained to a remarkably limited set of options. The present study leverages this tool-use pattern to connect an LLM with massive tools and retain token efficiency.




\textit{\textbf{Question 2.} How can an LLM identify ineffective tools and modify its tool-utilization strategies for future tasks?}

\cite{shinn2023reflexion} shows that LLMs can verbally reflect their tool-use trajectories for improvements within ongoing interactions or episodes. In this work, we consider fine-grained reflection at the tool level. Every tool-use step is probed and scored by an LLM. Scores will be used to adjust the tool transition weights, where low-quality tools will be restricted with reduced transition weights. Notably, ~\cite{shao2023prompting} shows that providing an LLM with scores of candidate answers brings significant performance improvements. Likewise, we prompt an LLM with the transition weights of tools. In this way, by encoding the experience into a tool graph, failure calls to low-quality tools can be significantly reduced.

To this end, this paper presents \sysname, a simple yet effective paradigm that assists LLMs in handling massive tools, learning to select appropriate tools, and avoiding calls to broken tools. A comparison between \sysname and other methods is provided in Table~\ref{table:comparison_with_sota}. As is shown in Figure~\ref{fig:react_agentnet_overall}, \sysname organizes the tools in a weighted directed graph. The graph is built based on the tool-use trajectories produced by an LLM. In turn, the LLM uses the graph to reason its tool calls, a process that can be conceptualized as navigating within this graph. 
Furthermore, the graph can be online updated, thereby enabling its adjustment to accommodate the frequent updates of tools or the introduction of new tasks. 
Extensive experiments are conducted on five distinct datasets: SciQA~\cite{lu2022learn}, TabMWP~\cite{lu2022dynamic}, MATH~\cite{hendrycksmath2021}, APIBank~\cite{li2023api}, and ToolBench~\cite{qin2023toolllm}. The results indicate that \sysname consistently outperforms its counterparts in terms of overall performance. Notably, it demonstrates remarkable resilience against the interference of noisy tools and achieves this superior performance while utilizing significantly fewer tokens.


\begin{table}[t]
	\centering
    \caption{
    A comparison of work that augments LLMs with tool usage. \textit{adapt.} denotes the adaptability to new tools or tools with updates. NL denotes natural language. Plug-Play means the LLMs can be equipped and unequipped with a tool flexibly.}
    \label{table:comparison_with_sota}
    \resizebox{0.48\textwidth}{!}{
        \begin{tabular}{l|ccc|cc}
\toprule
\multirow{2}{*}{Method} & \multicolumn{3}{c|}{API/Tool Use}                                                                        & \multicolumn{2}{c}{Framework}                                                                                               \\ \cline{2-6} 
                        & \#APIs & Adapt. & \begin{tabular}[c]{@{}c@{}}Token\\ Efficiency\end{tabular} &  Planning & \begin{tabular}[c]{@{}l@{}}Plug.\\ Play\end{tabular} \\ \midrule
Toolformer              & 5      & --               & +                                                                                                    & -        & \ding{55}                           \\
ReAct                   & 3      & +                & -                                                                                                                           & NL       & \ding{51}                           \\
ToT                     & -      & --               & --                                                                                                                         & NL       & \ding{51}                           \\
Reflexion               & -      & +                & --                                                                                                                        & NL       & \ding{51}                           \\
Chameleon               & 15     & -                & -                                                                                                      & NL       & \ding{51}                           \\
HuggingGPT              & 24     & +                & -                                                                                                      & NL       & \ding{51}                           \\
Gorilla                 & 1645   & -                & +                                                                                                    & -        & \ding{55}                           \\
RestGPT                 & 100+   & +                & -                                                                                                       & NL       & \ding{51}                           \\ \midrule
\sysname (\textbf{ours})          & 3992   & ++               & +                                                                                                                             & NL+Graph & \ding{51}                           \\ \bottomrule
\end{tabular}
    }
\end{table}

%% file: sections/background.tex
\label{section:background}
Tool-augmented LLMs interact with the environment through structured texts in natural language. The general interaction process can be described as follows: at step $s$, an LLM takes the current environmental observation $o_s$ along with a history of interaction activities $\mathcal{C}_s$ as the input and produces a verbal thought $t_s \in \mathcal{L}$. Then, the LLM interacts with the environment by taking an action $a_s \in \mathcal{A}$, which can be described as a tuple of a tool name and its arguments, e.g., $a_s$ = $(tool_s, arg_s)$, where $tool_s$ $\in$ $\mathcal{T}$. Subsequently, the LLM gets a new observation $o_{s+1}$ from the environment $\mathcal{E}$. 
By iterating the interaction process, the LLM may eventually resolve a given task.


LLMs are mostly stateless. The history of interactions $\mathcal{C}_s$ is generally formulated as part of the input to LLMs. It helps an LLM recall the previous process and infer the next action $a_{s+1}$. We define $\mathcal{C}_s$ as a queue of size $K$, consisting of observations, thoughts and actions in the past, e.g., $\mathcal{C}_s$ = $[(o_{s-k}, t_{s-k}, a_{s-k}), \cdots, (o_{s-1}, t_{s-1}, a_{s-1})]$. In each step, a tuple of new thought, action, and observation is appended to $\mathcal{C}_s$, and if the maximum capacity is reached, the earliest tuple is popped out. In addition to $\mathcal{C}_s$, the set of available tools $\mathcal{T}$ needs to be included as the input context as well, from which, the LLM may choose a proper one.



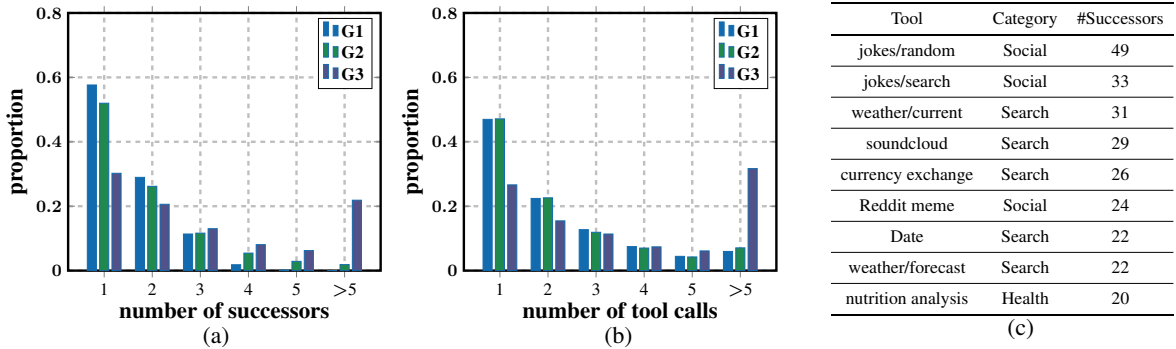
\begin{figure*}[!t]
\hspace{10px}
\begin{minipage}[t]{0.32\textwidth}
\captionsetup[subfloat]{captionskip=0pt, labelformat=empty}
\subfloat[\hspace{1em} \hspace{-12pt}(a)]
{
    \label{fig:toolbench_data_proportion}
    \input{Fig/toolbench_data}
}
\end{minipage}
\hspace{-10px}
\begin{minipage}[t]{0.32\textwidth}
\captionsetup[subfloat]{captionskip=0pt, labelformat=empty}
\subfloat[\hspace{1em} (b)]
{
    \label{fig:toolbench_data_count}
    \input{Fig/toolbench_count}
}
\end{minipage}
\hspace{2px}
\begin{minipage}[t]{0.32\textwidth}
\captionsetup[subfloat]{captionskip=0pt, labelformat=empty}
\subfloat[\hspace{1em} (c)]
{
    \label{fig:toolbench_data_overall}
    \input{Fig/toolbench_overall}
}
\end{minipage}
\vspace{-8pt}
\caption{The tool-use statistics on Toolbench, which consists of three subsets: G1, G2, and G3. (a) shows that around 80\% of tools have less than 6 successor tools called. (b) illustrates that over 90\% of tools in the G1 and G2 subsets, as well as over 50\% of tools in the G3 subset, are called fewer than 6 times. (c) lists tools with the most successor tools in the G3 subset. These statistics motivate the proposed \sysname to construct a tool graph with sparse connections of tools.}
\label{fig:toolbench_data}
\end{figure*}

Let $\mathcal{P}_\theta$ denotes a tool-augmented LLM with parameters $\theta$. Formally, the interaction process described above can be formulated as follows.
\begin{align}
    t_s &=\mathcal{P}_\theta(\mathcal{C}_s\cup o_s)\label{eq:thought},\\
    a_s &=\mathcal{P}_\theta(\mathcal{C}_s\cup o_s, t_s, \mathcal{T})\label{eq:action},\\
    o_{s+1} &= \mathcal{E}(a_s)
\end{align}
A predominant issue of existing tool-augmented LLMs is their heavy token consumption, which can be largely attributed to the long input context of $\mathcal{C}_s$ and $\mathcal{T}$. A representative solution is by segmenting the long texts of $\mathcal{C}_s$ into paragraphs and compressing them into semantic embeddings. Subsequently, the semantic similarity scores between the candidate paragraphs and the current observation $o_s$ can be computed and ranked. Finally, $\mathcal{C}_s$ in Eq.~\ref{eq:thought}-~\ref{eq:action} is replaced by the most relevant paragraphs as the input to LLMs. Intuitively, this solution can be extended to $\mathcal{T}$, e.g., searching tools in $\mathcal{T}$ that are the most relevant to the thought $t_s$. However, according to Eq.~\ref{eq:thought}, $t_s$ is generated without the full knowledge of $\mathcal{T}$. The semantic information in $t_s$ can be irrelevant to any tools in $\mathcal{T}$. Moreover, in a library of massive API functions, there can be many tools sharing similar semantic information but differing subtly in functionalities. Searching tools by semantic embeddings can be coarse and inaccurate when $|\mathcal{T}|$ goes large. 
Consequently, existing tool learning approaches either are restricted to a limited number of specifically designed tools or use instruction finetuning to augment the LLM $\mathcal{P}_\theta$ with the prior knowledge of $\mathcal{T}$, which can be computationally intensive.

%% file: Fig/toolbench_data.tex
\begin{tikzpicture}[scale=0.6,trim left=-0.3cm,  baseline=-0.515cm]
\begin{axis}[
    ybar,
    axis line style={line width=1.5pt},
    label style={font=\Large},
    xlabel={\textbf{number of successors}},
    ylabel={\textbf{proportion}},
    xticklabels={1,2,3,4,5,$>$5},
    xtick={1,2,3,4,5,6},
    ymin=0.0, ymax=0.8,
    enlargelimits=0.15,
    enlarge y limits=false,
    grid=both,          
    grid style={dashed, line width=1.5pt},  
    y label style={yshift=-3pt}, 
    x label style={yshift=-1pt}, 
    legend pos=north west,
    tick label style={font=\boldmath},
    bar width=0.20cm,
    legend style={at={(0.80,0.98)},anchor=north west,cells={anchor=west}},
]
\definecolor{color1}{RGB}{18,107,174}
\definecolor{color2}{RGB}{32,137,77}
\definecolor{color3}{RGB}{82,82,136}
\addplot[draw=color1, fill=color1] coordinates{
(1, 0.5764028056112225)
(2, 0.2895791583166333)
(3, 0.11372745490981964)
(4, 0.018286573146292586)
(5, 0.0017535070140280561)
(6, 0.000250501002004008)
};
\addlegendentry{\textbf{G1}}
\addplot[draw=color1, fill=color2] coordinates{
(1, 0.5198049969530774)
(2, 0.261832216128377)
(3, 0.11639244363193176)
(4, 0.054438350599228115)
(5, 0.028844200690635792)
(6, 0.01868779199674995)
};
\addlegendentry{\textbf{G2}}
\addplot[draw=color1, fill=color3] coordinates{
(1, 0.3022351797862002)
(2, 0.20602526724975703)
(3, 0.13022351797862003)
(4, 0.08066083576287658)
(5, 0.062196307094266275)
(6, 0.21865889212827988)
};
\addlegendentry{\textbf{G3}}
\end{axis}
\end{tikzpicture}

%% file: Fig/toolbench_count.tex
\begin{tikzpicture}[scale=0.6, trim left = -0.55cm, baseline=-0.515cm]
\begin{axis}[
    ybar,
    axis line style={line width=1.5pt},
    label style={font=\Large},
    xlabel={\textbf{number of tool calls}},
    ylabel={\textbf{proportion}},
    xticklabels={1,2,3,4,5,$>$5},
    xtick={1,2,3,4,5,6},
    ymin=0.0, ymax=0.8,
    enlargelimits=0.15,
    enlarge y limits=false,
    grid=both,          
    grid style={dashed, line width=1.5pt},  
    y label style={yshift=-3pt}, 
    x label style={yshift=-1pt}, 
    legend pos=north west,
    tick label style={font=\boldmath},
    bar width=0.20cm,
    legend style={at={(0.80,0.98)},anchor=north west,cells={anchor=west}},
]
\definecolor{color1}{RGB}{18,107,174}
\definecolor{color2}{RGB}{32,137,77}
\definecolor{color3}{RGB}{82,82,136}
\addplot[draw=color1, fill=color1] coordinates{
(1, 0.46968937875751504)
(2, 0.22419839679358716)
(3, 0.12725450901803606)
(4, 0.0746492985971944)
(5, 0.04433867735470942)
(6, 0.05961923847695391)
};
\addlegendentry{\textbf{G1}}
\addplot[draw=color1, fill=color2] coordinates{
(1, 0.4712573633963031)
(2, 0.22608165752589884)
(3, 0.11903310989234207)
(4, 0.07007921998781232)
(5, 0.04245378834044282)
(6, 0.07068860450944547)
};
\addlegendentry{\textbf{G2}}
\addplot[draw=color1, fill=color3] coordinates{
(1, 0.2662779397473275)
(2, 0.15451895043731778)
(3, 0.11370262390670553)
(4, 0.0738581146744412)
(5, 0.061224489795918366)
(6, 0.3168124392614186)
};
\addlegendentry{\textbf{G3}}
\end{axis}
\end{tikzpicture}

%% file: Fig/toolbench_overall.tex
\resizebox{0.88\linewidth}{!}{\begin{tabular}{ccc}
        \toprule[1.2pt]
        Tool & Category &  \#Successors\\
        \midrule[1.2pt]
        jokes/random & Social & 49\\
         \midrule
        jokes/search & Social & 33  \\
         \midrule
        weather/current & Search & 31 \\
         \midrule
        soundcloud & Search & 29  \\
         \midrule
        currency exchange & Search & 26 \\
        \midrule
        Reddit meme & Social & 24\\
        \midrule
        Date & Search & 22 \\
        \midrule
        weather/forecast & Search & 22 \\
        \midrule
        nutrition analysis & Health & 20 \\
        \bottomrule[1.2pt]
\end{tabular}}

%% file: sections/method.tex
\label{section:toolnet}
The idea of \sysname is simple. Instead of inputting all tools in $\mathcal{T}$ to $\mathcal{P}_\theta$ as in Eq.~\ref{eq:action}, we provide only a subset $\mathcal{T}_s \subset \mathcal{T}$, according to a policy $\pi$ and the previous action $a_{s-1}$, i.e.,
\begin{align}
    a_s &= \mathcal{P}_\theta(\mathcal{C}_s \cup o_s, t_s, \mathcal{T}_s),
\end{align}
where $\mathcal{T}_s$ = $\pi(\mathcal{T}, a_{s-1})$. Markov assumption takes effect in the generation of $\mathcal{T}_s$. We describe $\pi$ as a weighted directed graph $\mathcal{G}$ = $(\mathcal{V}, \mathcal{E})$ that connects all the tools in $\mathcal{T}$. $\mathcal{V}$ and $\mathcal{E}$ denote the sets of nodes and edges, respectively. 
Every tool in $\mathcal{T}$ is regarded as a unique node in $\mathcal{V}$. 
In addition, two special nodes, $start$ and $end$, are added into $\mathcal{V}$. In other words, $\mathcal{V}$ = $\mathcal{T}\cup\{start, end\}$. An edge $e$ = $(v_i, v_j, w_{i,j})$ connects node $v_i$ to node $v_j$, with $w_{i,j}$ denotes the transition weight. Note that tools along with the transition weights are formatted as the input texts to LLMs. All nodes except $start$ and $end$ have a self-transition edge that directs to themselves. The $start$ node directs to all the other nodes, which in turn directs to the $end$.

To this end, an LLM taking actions is the same as traveling in the graph $\mathcal{G}$. Assume the LLM used $tool_{s-1}$ = $v_i$ in the last iteration. The available tools $\mathcal{T}_s$ in the next step are simply the out-neighbor nodes of $v_i$, e.g., $\mathcal{T}_s$ = $oneigh(v_i)$. If the graph $\mathcal{G}$ is sparse, $|\mathcal{T}_s| \ll |\mathcal{T}|$, bringing token efficiency to tool-augmented LLMs.
Notably, the $start$ node connects all tools. In other words, the LLM starts with $\mathcal{T}_1$ = $\mathcal{T}$, which bottlenecks the token consumption and becomes impractical as $|\mathcal{T}|$ scales up. To handle this problem, we leverage the semantic similarity search approach mentioned in Section~\ref{section:background}. The description of each tool is compressed into a semantic embedding, which is then ranked based on its similarity to the embedding of the initial task description $o_1$. Only the $K$ most relevant tools are kept to set up $\mathcal{T}_1$.


An additional benefit introduced by \sysname is the transition weights of tools in $\mathcal{T}_s$, which measure the prior preference of tools. Based on the weights, tools can be sorted in order and formatted as the input context. The LLM can refer to the preference scores to make selections. In contrast, the existing approaches, such as ReAct and Reflexion, provide an LLM with tools of equal preference. The LLM relies solely on its internal reasoning ability to make a choice.

\subsection{Graph Construction}
\label{section:graph_construction}


The construction of the graph $\mathcal{G}$ is vital to the performance and efficiency. If all tools in $\mathcal{G}$ are mutually connected with equal edge weights, \sysname becomes uninformative and degrades into the naive ReAct approach. A good graph $\mathcal{G}$ should be sparse, i.e., nodes have a small number of out-neighbors. Out-of-date tools that are no longer functional should be downweighted. Notably, graph construction can be flexible. This paper considers static and dynamic graph construction approaches. Other graph operations, such as composition, pruning, and partition, are worthy of exploration but beyond the scope of this paper.

\subsection{Static Construction}
\label{section:static_construction}
Static construction requires large amounts of tool-use trajectories, such as the massive handwritten code snippets on GitHub that call PyTorch API functions. The orders of API function calls can be leveraged to build up $\mathcal{G}$. Tool-use trajectories can also be generated by tool-augmented LLMs. The multi-step reasoning trajectories represent sequences of tool calls. Several existing works such as ToolBench~\cite{qin2023toolllm}, APIBench~\cite{peng2022revisiting}, API-Bank~\cite{li2023api}, and ToolAlpaca~\cite{tang2306toolalpaca}, have released their curated tool-use instances, from which \sysname can be constructed. Unfortunately, the quality of trajectories generated from LLMs is of great concern.


The static construction of $\mathcal{G}$ is straightforward. It is the same as 2-gram language modeling. Denote $\mathcal{D}$ as a set of tool-use trajectories that have completed their tasks.  A trajectory has a sequence of tool uses, e.g., $[tool_1, \cdots, tool_s, \cdots, end]$, where, for simplicity, we regard $end$ as a normal tool.
The transition weight $w_{i,j}$ from node $v_i$ to node $v_j$ is computed as follows,
\begin{equation}
    w_{i,j} = \frac{\mathbb{E}_D\left[\mathds{1}\left(tool_s =v_i, tool_{s+1}=v_j\right)\right]}{\mathbb{E}_D\left[\mathds{1}\left(tool_s =v_i\right)\right]}
\end{equation}
The $start$ node in $\mathcal{G}$ is set up separately. It connects all tools except the $end$ with equal weights. When tools become massive, a tool retriever based on semantic similarity search can be adopted to select the first one, $tool_1$, as mentioned in Section~\ref{section:toolnet}.

\subsection{Dynamic Construction}
\label{section:dynamic_construction}
Tools are dynamically changing. They have life cycles and may no longer be maintained by developers. Consequently, the edge weights of $\mathcal{G}$ need timely finetuning, making static construction inapplicable. Moreover, there is generally a lack of large amounts of tool-use trajectories, especially for the emergent plugins on ChatGPT. In this case, dynamic construction of $\mathcal{G}$ is demanding.

$\mathcal{G}$ can be initialized either by static construction or as a non-informative graph with fully connected tools. 
Then, dynamic construction iterates between generating tool-use trajectories and updating $\mathcal{G}$. The process of trajectory generation is the same as explained in Section~\ref{section:toolnet}. The update of $\mathcal{G}$ is important. Since the number of tool-use trajectories is limited, a fine-grained inspection of trajectories is needed. An LLM acts as a tool evaluator, taking the whole trajectory as input and scoring every tool used. 
These scores are discrete integers in $[-3, 3]$. Evaluating the tools equates to assessing the nodes visited in $\mathcal{G}$.
We use $\Delta_i^{(n)}$ to denote the Evaluator's score of node $v_i$ in the $n$-th iteration. Let $s_i^{(n)}$ denote the accumulated score, e.g., $s_i^{(n)}$ = $\sum_{k=1}^n \Delta_i^{(k)}$. The transition weight $w_{i,j}^{(n)}$ is updated as follows.
\begin{align}
    w_{i,j}^{(n)} &= \beta w_{i,j}^{(0)} + (1-\beta)\Delta w_{i,j}^{(n)},\\
    \Delta w_{i,j}^{(n)} &\coloneqq \frac{f(s_i^{(n)})}{\sum_{v_j\in oneigh(v_i)} f(s_j^{(n)})},\\
    f(x) &\coloneqq \left\{
                \begin{array}{lll}
                  \alpha x+1, & x\geq 0\\
                  e^{\alpha x}, & x < 0
                \end{array}
              \right. 
,\end{align}
where $f(x)$ maps accumulated scores to $(0, +\infty)$. $\Delta w_{i,j}^{(n)}$ $\in$ $(0, 1]$ denotes the normalized gradient to update the transition weight. $\alpha$ and $\beta$ are hyper-parameters. $\alpha$ controls the speed of updating. $\beta$ interpolates between the prior weight $w_{i,j}^{(0)}$ and the weight learnt from dynamic construction.

%% file: sections/exp.tex
\label{section:experiments}
\subsection{Setup}

\textbf{Datasets.}  We conduct experiments on five datasets: (1) \textbf{SciQA}~\cite{lu2022learn}, a question-answering benchmark in various scientific fields including multiple subjects like biology, geography, and ecosystems. (2) \textbf{TabMWP}~\cite{lu2022dynamic}, a question-answering benchmark, where an LLM should answer questions according to a given Table. (3) \textbf{MATH}~\cite{hendrycksmath2021}, a question answering benchmark mathematical fields including 7 categories: Prealgebra,
Algebra, Number Theory, Counting and Probability, Geometry, Intermediate Algebra, and Precalculus. (4) \textbf{APIBank}~\cite{li2023api}, a multi-task benchmark consisting of various tools to evaluate the performance of tool-augmented LLMs.
(5) \textbf{ToolBench}~\cite{qin2023toolllm}, a large-scale benchmark with 3451 APIs spanning distinct domains.
Due to cost considerations, we select subsets of these datasets for evaluation. 
From the SciQA dataset, we randomly select 1,000 questions for evaluation.
For TabMWP, we utilize the test1k subset.
Within the MATH dataset, we exclusively consider level-5 questions, which represent the highest difficulty level. Additionally, we exclude geometry-related questions, aligning with prior research~\cite{drori2022neural, wu2023empirical}.
For APIBank, we focus on a specific subset, specifically the test data falling within the lv1-lv2-samples category, totaling 212 questions. In the ToolBench dataset, we select a random subset of 1,000 questions from the G1 set.

\textbf{Baselines.} 
(1) \textbf{ReAct}~\cite{yao2022react}. Equipped with various tools, an LLM strategically chooses one tool at a time for each step to address specific tasks or questions, ensuring a step-by-step, focused approach to problem-solving.
(2) \textbf{Reflexion}~\cite{shinn2023reflexion}). Similar to ReAct, LLM interacts with multiple tools in several steps. When an LLM fails a task, it will be prompted to verbally reflect on its reasoning trajectories. The self-refined LLM can learn from the prior errors and retry until the correct answer is submitted or the maximum tries are reached. (3) \textbf{Tree-of-Thought} (ToT)~\cite{yao2023tree}. It formulates the LLM reasoning process as a tree of thoughts,  
where each node within the tree symbolizes a segment of the solution. At any given node, a thought generator module is responsible for the creation of new nodes, effectively branching out the solution space. Subsequently, each of these new nodes undergoes evaluation. 
The progression and expansion of this thought tree are governed by a specific search algorithm, such as depth-first search (DFS), which dictates the sequence and manner the nodes are extended and explored.
\textbf{Implementation details.} We use \textit{gpt-3.5-turbo} as the LLM in all our experiments. Given a question or task instruction, the LLM iteratively selects one from the available tools till the \textit{Answer/Finish} tool is invoked to submit their answer or claim task failure when a maximum iteration of 8 is reached. 


\textbf{Metrics.} Evaluation is carried out in the aspects of answer quality and token consumption. Exact match (EM) is adopted to compare the answers generated from LLMs with the groundtruth provided in SciQA, TabMWP, MATH, and APIBank. There is no unique groundtruth answer in ToolBench, \textit{win rate} is therefore adopted. It measures the proportion of answers identified as correct by an external LLM-powered evaluator. 
In ToolBench, the built-in ToolEval (powered by \textit{gpt-3.5-turbo}) is adopted to compute the win rate. In addition, we measure the number of tokens consumed by LLMs for task completion. It takes into consideration both the input prompts and the generated tokens for reasoning and answering. 

\input{Fig/three_datasets_tools}
\subsection{Results and Analysis}

\subsubsection{Evaluation on Task-Specific Datasets}
In practice, noisy and task-irrelevant tools will inevitably exist when LLMs are connected with massive tools. To simulate this scenario, we first carried out experiments on three task-specific QA datasets, namely, SciQA, TabMWP, and MATH. The LLM, \textit{gpt-3.5-turbo}, is connected with 16 tools, of which most are irrelevant to the task the dataset was designed for. Table~\ref{table:three_datasets_tools} lists some typical tools used.
To the best of our knowledge, there are no public-available tool-use trajectories on these datasets. Therefore, \sysname creates the tool graph by dynamic construction, starting with a fully-connected graph with equal edge scores. 

Table~\ref{table:result_of_toy_datasets} compares existing tool learning methods (ReAct, Relfexion, and ToT) with the proposed \sysname. 
Compared with Reflexion, \sysname uses only $16.4\%$ and $22.8\%$ tokens on TabMWP and MATH, respectively. In terms of answer quality (measured by EM), \sysname significantly surpasses all the other competing approaches on SciQA and TabMWP, achieving a minimum absolute improvement of $10\%$. Admittedly, Reflexion is slightly better than \sysname on MATH but suffers from significantly increased token consumption and more reasoning steps. 
In addition, we study the effect of noisy tools on tool-augmented LLMs, with ReAct (clean) as a showcase. It denotes running ReAct by removing all noisy tools on the three datasets. Compared to ReAct (clean), ReAct suffers from dramatic answer quality degradation, increased token consumption, and longer reasoning steps, suggesting its inefficacy in handling noisy and task-irrelevant tools.

\begin{figure}[h!]
\centering
        \includegraphics[width=0.48\textwidth]{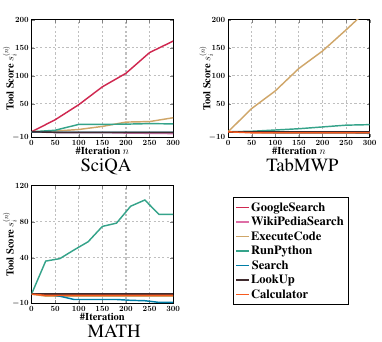} 
        \vspace{-2em}
        \caption{
Analysis of tool scores on SciQA, TabMWP, and MATH in dynamic graph construction. Tools effective for each task-specific dataset exhibit increasing scores, indicating \sysname's capability to automatically select the most appropriate tools for each specific task.}
        \label{fig:result_of_toy_datasets}
\end{figure}

In our analysis of the tool scores $s_i^{(n)}$ generated by \sysname, as illustrated in Figure~\ref{fig:result_of_toy_datasets}, we observe a clear pattern. For each task-specific dataset, the scores of effective tools show a consistent increase, whereas the scores for irrelevant tools remain approximately zero. This pattern demonstrates \sysname's proficiency in selecting the most suitable tools for distinct tasks. Specifically, \textit{GoogleSearch} emerges as the preferred tool for the SciQA dataset. In contrast, for the TabMWP and MATH datasets, \textit{ExecuteCode} and \textit{RunPython} are identified as the most appropriate tools, respectively. Additionally, \sysname effectively down-weights noisy tools, further optimizing the tool selection process.


\input{Fig/result_of_toy_datasets}

\subsubsection{Evaluation on Multi-Task Datasets with Massive Tools}
We then evaluate \sysname on APIBank and ToolBench, two challenging datasets that consist of diverse tasks and have significantly large amounts of available tools. The massive tools in ToolBench make LLMs difficult to select the starting tool. We follow the default setup in ToolBench to use Tool Retriever, a fine-tuned BERT model, to recommend the $K$ most proper starting tools to LLM. We set $K$ = $8$ to augment ReAct and Reflexion on ToolBench and $K$ = $5$ on APIBank. For \sysname and \textit{ToT}, only the most recommended tool is provided as the starting tool. 
The graph of tools is firstly statically constructed from the trajectories provided in the datasets, as is explained in Section~\ref{section:static_construction}. Thereafter, the graph is updated dynamically at every iteration.


As indicated in Table ~\ref{table:result_of_large_datasets}, \sysname is on par with Reflexion in terms of win rate, while utilizing significantly fewer tokens (only 38.5\% on APIBank and 49.7\% on ToolBench, compared to Reflexion). Additionally, Reflexion's methodology is complementary to \sysname. When integrated into \sysname, it further enhances answer quality, leading to the highest win rates on both datasets.

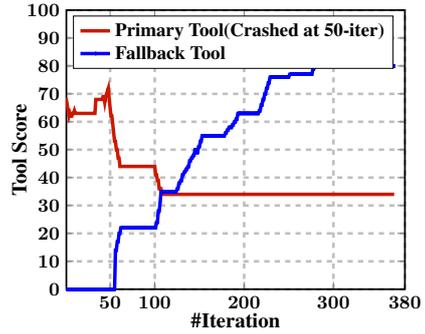
\begin{figure}[htbp]
    \centering
    \input{Fig/dyn_toolfail}
    \caption{
    Scores of primary and fallback tools with identical functionality (\textit{GetUserToken} of APIBank) in dynamic graph construction. The primary tool crashed at the 50th iteration. \sysname effectively down-weighs the crashed tool and revives the fallback one, demonstrating its resilience against sudden tool failures.}
    \label{fig:dynamic_tool_failure}
\end{figure}

The large number of tools and the diverse tasks provided by the two datasets prohibit an intuitive interpretation of tool scores and manual tool selection by humans. Meanwhile, tools are alive and constantly evolving. Some tools may occasionally get crashed and can no longer used. 
This demands \sysname to adaptively raise the sores of some other backup tools. We showcase that dynamic construction can mitigate such a problem. In Figure~\ref{fig:dynamic_tool_failure}, we simulate on APIBank the scenario where a tool suddenly breaks at the 50th iteration. The score of the broken tool experiences a significant decline whereas the backup tool is swiftly activated and effectively supplants the impaired one. This transition underscores the robust adaptability inherent in \sysname. In this regard, tool redundancy serves as a critical feature in enhancing system reliability.

\input{Fig/result_of_large_datasets}

\subsubsection{Effect of Transition Weights}
Tools along with the transition weights are formatted as the input contexts to LLMs. The transition weights indicate the priorities of tools, thus playing a vital role in tool selection. This section analyses how the transition weights affect the overall performance. As shown in Table \ref{tab:edge_weight_scaling}, five test conditions are considered: removing weights from the input to LLMs (No weight), weights divided by 100 with decimals kept ($/100$), weights divided by 10 with decimals kept ($/10$), removing decimal parts of weights (Integer), and multiplying weights by 10 (×10). Experiments in each condition were repeated three times to reduce randomness.
\begin{table}[t]
\centering
\resizebox{0.48\textwidth}{!}{ 
\begin{tabular}{l|ccccc}
\toprule
 & \multicolumn{5}{c}{Modification of Transition Weights} \\
 \cline{2-6}
Metric & No weight & $/100$ & $/10$ & Integer & $\times10$\\ \midrule
EM & 0.70 & 0.70 & 0.71 & \textbf{0.75} & \textbf{0.75}\\
\#Tokens & 1710 & 1682 & 1664 & 1649 & \textbf{1645}\\
\bottomrule
\end{tabular}
}
\caption{How transition weights affect the performance of \sysname. The evaluation is carried out on APIBank.}
\label{tab:edge_weight_scaling}
\end{table}
Worst performance is attained when the weights are removed or scaled down into decimals. This suggests the importance of a proper format of weights. Moreover, there is a general increase in EM when scaling up the transition weights. This suggests that LLMs are more sensitive to the difference between integers or large numbers than the subtle difference between decimals. The token consumption is also gradually reduced when scaling up the weights. These observations may indicate a potential optimization pathway for tool learning. Notably, there is an upper-performance limit when scaling up the weights, e.g., the performance saturates at $\times10$.

\input{Fig/result_of_toolbench}

%% file: Fig/three_datasets_tools.tex
\begin{table}[t]
	\renewcommand{\arraystretch}{1.3}
	\centering
    \caption{
    A list of representative tools utilized in three task-specific QA datasets. \ding{55} denotes tools are irrelevant or noisy to the dataset.
    }
    \label{table:three_datasets_tools}
    \resizebox{0.5\textwidth}{!}{ 
        \begin{tabular}{C{2.3cm}C{4cm}C{0.8cm}C{1.2cm}C{1cm}}
            \toprule[1.2pt]
            \multirow{2}{*}{Tool} & \multirow{2}{*}{Description} & \multicolumn{3}{c}{Dataset} \\
            \cline{3-5} 
               & & SciQA & TabMWP & MATH \\ 
            \midrule[1.2pt]
            GoogleSearch & Search by google & \ding{51} & \ding{55} & \ding{55}  \\
            WikiPediaSearch & Search in Wikipedia & \ding{51} & \ding{55} & \ding{55} \\
            ExecuteCode & Execute math expressions & \ding{55} & \ding{51} & \ding{51} \\
            RunPython & Run python code &  \ding{55} & \ding{51} & \ding{51} \\
            Search$^{\mathrm{*}}$ & Return 'Nothing Found' & \ding{55} & \ding{55} & \ding{55} \\
            LoopUp$^{\mathrm{*}}$ & Return 'Nothing Found' & \ding{55} & \ding{55} & \ding{55} \\
            Calculator$^{\mathrm{*}}$ & Return random numbers & \ding{55} & \ding{55} & \ding{55} \\ 
            \bottomrule[1.2pt]
            \multicolumn{4}{l}{$^{\mathrm{*}}$ means this tool is on purpose designed to be noisy.}
        \end{tabular}
    }
\end{table}

%% file: Fig/result_of_toy_datasets.tex
\begin{table}[t]

	\renewcommand{\arraystretch}{1.3}
	\centering
    \caption{Results on subsets of close-set task-specific datasets. Noisy and task-irrelevant tools, as described in Table.\ref{table:three_datasets_tools}, are introduced to analyze the robustness of prompting paradigms. Reflexion is not evaluated on SciQA, a multiple-choice QA dataset, as it would not provide a fair comparison.}
    \label{table:result_of_toy_datasets}
    \resizebox{0.48\textwidth}{!}{ 
\begin{tabular}{C{1.2cm}C{0.5cm}c|cccccg}
\toprule
Dataset & Size               & \#APIs             & Metric  & ReAct & \begin{tabular}[c]{@{}c@{}}ReAct\\ ({clean})\end{tabular} & Reflexion                      & ToT   & \begin{tabular}[c]{@{}c@{}}ToolNet\\ (\textbf{Ours})\end{tabular}                 \\ \hline
\multirow{3}{*}{SciQA} & \multirow{3}{*}{1000} & \multirow{3}{*}{16} & EM      & 0.45 & 0.48 & -                              & 0.12  & \textbf{0.61} \\
                           &                       &                     & \#Tokens  & 8270  & 7318 & -                              & 23070 & \underline{5945}                           \\
                           &                       &                     & \#Steps  & 5.8  & 5.4 & -                              & 8.9   & 4.0                            \\ \hline
\multirow{3}{*}{TabMWP}  & \multirow{3}{*}{1000} & \multirow{3}{*}{16} & EM  & 0.26  & 0.50 & \underline{0.57}                           & 0.09  & \textbf{0.67} \\
                           &                       &                     & \#Tokens  & 8936 & 6594 & 24710                          & 19669 & \underline{4062}                           \\
                           &                       &                     & \#Steps  & 6.1  & 5.3 & 11.3$^{\mathrm{*}}$            & 8.7   & 3.7                            \\ \hline
\multirow{3}{*}{MATH}      & \multirow{3}{*}{675}  & \multirow{3}{*}{16} & EM     & 0.13 & 0.20  & \textbf{0.29} & 0.07  & \underline{0.25}                           \\
                           &                       &                     & \#Tokens  & 9216 & 8382 & 28886                          & 18090 & \underline{6602}                           \\
                           &                       &                     & \#Steps    & 6.9 & 6.8  & 9.5                            & 8.2   & 5.4           \\                 \bottomrule
\end{tabular}
    }
\end{table}

%% file: Fig/result_of_large_datasets.tex
\begin{table}[t]

	\renewcommand{\arraystretch}{1.3}
	\centering
    \caption{
    Results on multi-task datasets with massive tools. Following the prior work~\cite{qin2023toolllm}, a fine-tuned BERT model provided in ToolBench is leveraged to recommend tools for LLMs. ReAct, Reflexion, and ToT are recommended with 8 tools in every reasoning step. \textit{ToolNet} is provided only with one tool in the first step.}
    \label{table:result_of_large_datasets}
    \resizebox{0.48\textwidth}{!}{ 
\begin{tabular}{C{1.2cm}C{0.6cm}C{0.9cm}|cccC{0.8cm}gg}
\toprule
\multirow{2}{*}{Dataset}   & \multirow{2}{*}{Size} & \multirow{2}{*}{\#APIs} & \multirow{2}{*}{Metric} & \multirow{2}{*}{ReAct} & \multirow{2}{*}{Reflexion} & \multirow{2}{*}{ToT} & \multicolumn{2}{g}{ToolNet (Ours)}                                               \\ \cline{8-9} 
                           &                          &                          &                           &                        &                            &                      &  & +Reflexion \\ \midrule
\multirow{2}{*}{\small{APIBank}}   & \multirow{2}{*}{212}     & \multirow{2}{*}{49}      & EM                       & 0.70                  & \underline{0.77}                      & 0.63                & 0.75   & \textbf{0.83}                                                        \\
                           &                          &                          & \#Tokens                  & \underline{1720}                   & 4286                       & 2013                 & \textbf{1649}    & 3353                                                         \\ \hline
\multirow{2}{*}{\small{ToolBench}} & \multirow{2}{*}{1000}    & \multirow{2}{*}{3992}      & Win Rate                       & 0.61                  & \underline{0.71}                      & 0.60                & 0.69   &         \textbf{0.75}                                                     \\
                           &                          &                          & \#Tokens                  & \underline{8636}                   & 13217                      & 10167                & \textbf{6575}    &                            12548                                  \\ \bottomrule
\end{tabular}

    }
\end{table}

%% file: Fig/dyn_toolfail.tex
\begin{tikzpicture}[scale=0.65]
\begin{axis}[
    axis line style={line width=1pt},
    label style={font=\large},
    xlabel={\textbf{\#Iteration}},
    ylabel={\textbf{Tool Score}},
    xmin=1, xmax=380, 
    ymin=0, ymax=100,
    xtick={50, 100, 200, 300, 380},
    ytick={0,10,20,...,100},
    grid=both,          
    grid style={dashed, line width=1.5pt},  
    y label style={yshift=-8pt}, 
    x label style={yshift=5pt}, 
    legend pos=north west,
    tick label style={font=\boldmath},
    legend style={at={(0.02,0.98)},anchor=north west,cells={anchor=west}},
]
\definecolor{color1}{rgb}{0.094,0,1}
\addplot[
    color=color1,
    mark=square,
    mark size=0pt,
    line width=2pt
    ]
    coordinates {
(0, 68)
(1, 68)
(2, 66)
(3, 64)
(4, 62)
(5, 63)
(6, 62)
(7, 62)
(8, 63)
(9, 63)
(10, 64)
(11, 63)
(12, 63)
(13, 63)
(14, 63)
(15, 63)
(16, 63)
(17, 63)
(18, 63)
(19, 63)
(20, 63)
(21, 63)
(22, 63)
(23, 63)
(24, 63)
(25, 63)
(26, 63)
(27, 63)
(28, 63)
(29, 63)
(30, 63)
(31, 63)
(32, 63)
(33, 63)
(34, 68)
(35, 68)
(36, 68)
(37, 68)
(38, 68)
(39, 68)
(40, 68)
(41, 68)
(42, 69)
(43, 69)
(44, 67)
(45, 69)
(46, 70)
(47, 71)
(48, 72)
(49, 68)
(50, 65)
(51, 62)
(52, 62)
(53, 59)
(54, 56)
(55, 53)
(56, 53)
(57, 50)
(58, 50)
(59, 47)
(60, 47)
(61, 44)
(62, 44)
(63, 44)
(64, 44)
(65, 44)
(66, 44)
(67, 44)
(68, 44)
(69, 44)
(70, 44)
(71, 44)
(72, 44)
(73, 44)
(74, 44)
(75, 44)
(76, 44)
(77, 44)
(78, 44)
(79, 44)
(80, 44)
(81, 44)
(82, 44)
(83, 44)
(84, 44)
(85, 44)
(86, 44)
(87, 44)
(88, 44)
(89, 44)
(90, 44)
(91, 44)
(92, 44)
(93, 44)
(94, 44)
(95, 44)
(96, 44)
(97, 44)
(98, 44)
(99, 44)
(100, 44)
(101, 41)
(102, 41)
(103, 39)
(104, 39)
(105, 36)
(106, 36)
(107, 36)
(108, 34)
(109, 34)
(110, 34)
(111, 34)
(112, 34)
(113, 34)
(114, 34)
(115, 34)
(116, 34)
(117, 34)
(118, 34)
(119, 34)
(120, 34)
(121, 34)
(122, 34)
(123, 34)
(124, 34)
(125, 34)
(126, 34)
(127, 34)
(128, 34)
(129, 34)
(130, 34)
(131, 34)
(132, 34)
(133, 34)
(134, 34)
(135, 34)
(136, 34)
(137, 34)
(138, 34)
(139, 34)
(140, 34)
(141, 34)
(142, 34)
(143, 34)
(144, 34)
(145, 34)
(146, 34)
(147, 34)
(148, 34)
(149, 34)
(150, 34)
(151, 34)
(152, 34)
(153, 34)
(154, 34)
(155, 34)
(156, 34)
(157, 34)
(158, 34)
(159, 34)
(160, 34)
(161, 34)
(162, 34)
(163, 34)
(164, 34)
(165, 34)
(166, 34)
(167, 34)
(168, 34)
(169, 34)
(170, 34)
(171, 34)
(172, 34)
(173, 34)
(174, 34)
(175, 34)
(176, 34)
(177, 34)
(178, 34)
(179, 34)
(180, 34)
(181, 34)
(182, 34)
(183, 34)
(184, 34)
(185, 34)
(186, 34)
(187, 34)
(188, 34)
(189, 34)
(190, 34)
(191, 34)
(192, 34)
(193, 34)
(194, 34)
(195, 34)
(196, 34)
(197, 34)
(198, 34)
(199, 34)
(200, 34)
(201, 34)
(202, 34)
(203, 34)
(204, 34)
(205, 34)
(206, 34)
(207, 34)
(208, 34)
(209, 34)
(210, 34)
(211, 34)
(212, 34)
(213, 34)
(214, 34)
(215, 34)
(216, 34)
(217, 34)
(218, 34)
(219, 34)
(220, 34)
(221, 34)
(222, 34)
(223, 34)
(224, 34)
(225, 34)
(226, 34)
(227, 34)
(228, 34)
(229, 34)
(230, 34)
(231, 34)
(232, 34)
(233, 34)
(234, 34)
(235, 34)
(236, 34)
(237, 34)
(238, 34)
(239, 34)
(240, 34)
(241, 34)
(242, 34)
(243, 34)
(244, 34)
(245, 34)
(246, 34)
(247, 34)
(248, 34)
(249, 34)
(250, 34)
(251, 34)
(252, 34)
(253, 34)
(254, 34)
(255, 34)
(256, 34)
(257, 34)
(258, 34)
(259, 34)
(260, 34)
(261, 34)
(262, 34)
(263, 34)
(264, 34)
(265, 34)
(266, 34)
(267, 34)
(268, 34)
(269, 34)
(270, 34)
(271, 34)
(272, 34)
(273, 34)
(274, 34)
(275, 34)
(276, 34)
(277, 34)
(278, 34)
(279, 34)
(280, 34)
(281, 34)
(282, 34)
(283, 34)
(284, 34)
(285, 34)
(286, 34)
(287, 34)
(288, 34)
(289, 34)
(290, 34)
(291, 34)
(292, 34)
(293, 34)
(294, 34)
(295, 34)
(296, 34)
(297, 34)
(298, 34)
(299, 34)
(300, 34)
(301, 34)
(302, 34)
(303, 34)
(304, 34)
(305, 34)
(306, 34)
(307, 34)
(308, 34)
(309, 34)
(310, 34)
(311, 34)
(312, 34)
(313, 34)
(314, 34)
(315, 34)
(316, 34)
(317, 34)
(318, 34)
(319, 34)
(320, 34)
(321, 34)
(322, 34)
(323, 34)
(324, 34)
(325, 34)
(326, 34)
(327, 34)
(328, 34)
(329, 34)
(330, 34)
(331, 34)
(332, 34)
(333, 34)
(334, 34)
(335, 34)
(336, 34)
(337, 34)
(338, 34)
(339, 34)
(340, 34)
(341, 34)
(342, 34)
(343, 34)
(344, 34)
(345, 34)
(346, 34)
(347, 34)
(348, 34)
(349, 34)
(350, 34)
(351, 34)
(352, 34)
(353, 34)
(354, 34)
(355, 34)
(356, 34)
(357, 34)
(358, 34)
(359, 34)
(360, 34)
(361, 34)
(362, 34)
(363, 34)
(364, 34)
(365, 34)
(366, 34)
(367, 34)
(368, 34)

    };
    \addlegendentry{\textbf{Primary Tool(Crashed at 50-iter)}}
\definecolor{color1}{rgb}{0.8,0.094,0}

\addplot[
    color=blue,
    mark=asterisk,
    mark size=0pt,
    line width=2pt
    ]
    coordinates {
(0, 0)
(1, 0)
(2, 0)
(3, 0)
(4, 0)
(5, 0)
(6, 0)
(7, 0)
(8, 0)
(9, 0)
(10, 0)
(11, 0)
(12, 0)
(13, 0)
(14, 0)
(15, 0)
(16, 0)
(17, 0)
(18, 0)
(19, 0)
(20, 0)
(21, 0)
(22, 0)
(23, 0)
(24, 0)
(25, 0)
(26, 0)
(27, 0)
(28, 0)
(29, 0)
(30, 0)
(31, 0)
(32, 0)
(33, 0)
(34, 0)
(35, 0)
(36, 0)
(37, 0)
(38, 0)
(39, 0)
(40, 0)
(41, 0)
(42, 0)
(43, 0)
(44, 0)
(45, 0)
(46, 0)
(47, 0)
(48, 0)
(49, 0)
(50, 0)
(51, 0)
(52, 0)
(53, 0)
(54, 0)
(55, 0)
(56, 14)
(57, 14)
(58, 17)
(59, 17)
(60, 20)
(61, 20)
(62, 22)
(63, 22)
(64, 22)
(65, 22)
(66, 22)
(67, 22)
(68, 22)
(69, 22)
(70, 22)
(71, 22)
(72, 22)
(73, 22)
(74, 22)
(75, 22)
(76, 22)
(77, 22)
(78, 22)
(79, 22)
(80, 22)
(81, 22)
(82, 22)
(83, 22)
(84, 22)
(85, 22)
(86, 22)
(87, 22)
(88, 22)
(89, 22)
(90, 22)
(91, 22)
(92, 22)
(93, 22)
(94, 22)
(95, 22)
(96, 22)
(97, 22)
(98, 22)
(99, 22)
(100, 22)
(101, 22)
(102, 24)
(103, 24)
(104, 28)
(105, 28)
(106, 31)
(107, 35)
(108, 35)
(109, 35)
(110, 35)
(111, 35)
(112, 35)
(113, 35)
(114, 35)
(115, 35)
(116, 35)
(117, 35)
(118, 35)
(119, 35)
(120, 35)
(121, 35)
(122, 35)
(123, 35)
(124, 35)
(125, 36)
(126, 36)
(127, 38)
(128, 38)
(129, 39)
(130, 39)
(131, 41)
(132, 41)
(133, 42)
(134, 42)
(135, 43)
(136, 43)
(137, 44)
(138, 44)
(139, 46)
(140, 46)
(141, 47)
(142, 47)
(143, 48)
(144, 48)
(145, 50)
(146, 50)
(147, 49)
(148, 49)
(149, 51)
(150, 51)
(151, 53)
(152, 53)
(153, 55)
(154, 55)
(155, 55)
(156, 55)
(157, 55)
(158, 55)
(159, 55)
(160, 55)
(161, 55)
(162, 55)
(163, 55)
(164, 55)
(165, 55)
(166, 55)
(167, 55)
(168, 55)
(169, 55)
(170, 55)
(171, 55)
(172, 55)
(173, 55)
(174, 55)
(175, 55)
(176, 55)
(177, 55)
(178, 55)
(179, 56)
(180, 56)
(181, 57)
(182, 57)
(183, 58)
(184, 58)
(185, 58)
(186, 58)
(187, 59)
(188, 59)
(189, 60)
(190, 60)
(191, 61)
(192, 61)
(193, 63)
(194, 63)
(195, 63)
(196, 63)
(197, 63)
(198, 63)
(199, 63)
(200, 63)
(201, 63)
(202, 63)
(203, 63)
(204, 63)
(205, 63)
(206, 63)
(207, 63)
(208, 63)
(209, 63)
(210, 63)
(211, 63)
(212, 63)
(213, 63)
(214, 63)
(215, 63)
(216, 63)
(217, 64)
(218, 64)
(219, 66)
(220, 66)
(221, 68)
(222, 68)
(223, 70)
(224, 70)
(225, 72)
(226, 72)
(227, 74)
(228, 74)
(229, 76)
(230, 76)
(231, 76)
(232, 76)
(233, 76)
(234, 76)
(235, 76)
(236, 76)
(237, 76)
(238, 76)
(239, 76)
(240, 76)
(241, 76)
(242, 76)
(243, 76)
(244, 76)
(245, 76)
(246, 76)
(247, 76)
(248, 76)
(249, 76)
(250, 76)
(251, 77)
(252, 77)
(253, 77)
(254, 77)
(255, 77)
(256, 77)
(257, 77)
(258, 77)
(259, 77)
(260, 77)
(261, 77)
(262, 77)
(263, 77)
(264, 77)
(265, 77)
(266, 77)
(267, 77)
(268, 77)
(269, 77)
(270, 77)
(271, 77)
(272, 77)
(273, 77)
(274, 77)
(275, 77)
(276, 77)
(277, 79)
(278, 79)
(279, 81)
(280, 81)
(281, 81)
(282, 81)
(283, 81)
(284, 81)
(285, 81)
(286, 81)
(287, 81)
(288, 81)
(289, 81)
(290, 81)
(291, 81)
(292, 81)
(293, 81)
(294, 81)
(295, 81)
(296, 81)
(297, 81)
(298, 81)
(299, 81)
(300, 81)
(301, 81)
(302, 81)
(303, 81)
(304, 81)
(305, 81)
(306, 81)
(307, 81)
(308, 81)
(309, 81)
(310, 81)
(311, 81)
(312, 81)
(313, 81)
(314, 81)
(315, 81)
(316, 81)
(317, 81)
(318, 81)
(319, 80)
(320, 80)
(321, 80)
(322, 80)
(323, 80)
(324, 80)
(325, 80)
(326, 80)
(327, 80)
(328, 80)
(329, 80)
(330, 80)
(331, 80)
(332, 80)
(333, 80)
(334, 80)
(335, 80)
(336, 80)
(337, 80)
(338, 80)
(339, 80)
(340, 80)
(341, 80)
(342, 80)
(343, 80)
(344, 80)
(345, 80)
(346, 80)
(347, 80)
(348, 80)
(349, 80)
(350, 80)
(351, 80)
(352, 80)
(353, 80)
(354, 80)
(355, 80)
(356, 80)
(357, 80)
(358, 80)
(359, 80)
(360, 80)
(361, 80)
(362, 80)
(363, 80)
(364, 80)
(365, 80)
(366, 80)
(367, 80)
(368, 80)
    };
    \addlegendentry{\textbf{Fallback Tool}}


\end{axis}
\end{tikzpicture}

%% file: sections/related.tex
\textbf{Fine-tuning LLMs to use tools.}
Early research heavily relied on model fine-tuning for domain-specific tool learning and retrieval-augmented generation. 
Prominent works in this area include REALM~\cite{guu2020retrieval}, RETRO~\cite{borgeaud2022improving},VisualGPT~\cite{chen2022visualgpt}, etc. 
Notably, WebGPT~\cite{nakano2112webgpt}, a GPT-3 finetuned on human web search behaviors, can effectively utilize web browsers for question answering.
Furthermore, there has been a notable shift in research towards tuning LLMs on a broader spectrum of general tools. Example works include Toolformer~\cite{schick2023toolformer}, ToolkenGPT~\cite{hao2023toolkengpt}, Gorilla~\cite{patil2023gorilla}, ToolLLM~\cite{qin2023toolllm} etc. However, it is important to acknowledge that collecting tool-use data for fine-tuning can be prohibitively expensive, and these fine-tuned LLMs often struggle to generalize to emergent or updated tools.

\textbf{Prompting LLMs to use tools.}
The remarkable in-context learning ability of LLMs motivates prompt engineering approaches for tool learning~\cite{mialon2023augmented}. This is achieved by showing tool descriptions and demonstrations in context. Building on this idea, reasoning chains can be incorporated to tackle more complex problems, such as arithmetic calculation~\cite{cobbe2021training}, code execution~\cite{gao2023pal}, and complex mathematical theory verification~\cite{jiang2022draft}. More specifically, Plan-and-Solve~\cite{wang2023plan} enhances CoT with an explicit planning stage. Self-reflection~\cite{shinn2023reflexion,madaan2023self,paul2023refiner} and self-evaluation~\cite{xie2023decomposition} were introduced recently to self-correct mistakes in reasoning, showing enhanced performance in code generation and computer operation tasks.
These paradigms have given rise to popular industry products such as ChatGPT Store and LangChain~\cite{langchain2023}, as well as pioneering experimental products such as AutoGPT~\cite{richards2023auto}, MetaGPT~\cite{hong2023metagpt}, etc. Furthermore, by calling tools to interact with the virtual or physical world, LLMs are capable of guiding embodied agents to accomplish various household tasks~\cite{huang2022language,huang2022inner,singh2023progprompt}. Recent studies utilize LLMs as the central controller to coordinate multiple neural models, achieving promising results in multi-modal reasoning tasks~\cite{lu2023chameleon,shen2023hugginggpt}. Nevertheless, all methods based on in-context learning suffer from inferior performance in complex scenarios, where the tools are unfamiliar or numerous.


%% file: sections/conclusion.tex

We introduce \sysname, a novel plug-and-play method to organize massive tools into a directed graph, facilitating their use by LLMs via in-context learning. 
A key feature of \sysname is its adaptive tool transition weights, which can be initially set and continually updated. This adaptability ensures rapid integration of new tools and alignment with the changing environment of extensive tool repositories. 
Comprehensive evaluations on both task-specific datasets and complex, multi-task tool learning datasets have shown that \sysname consistently enhances performance and achieves up to $2.6$x greater token efficiency.
We believe that this research will inspire future studies to develop more advanced tool-augmented LLMs that can intelligently connect massive real-world tools to fulfill diverse human requirements.

%% file: sections/limitation.tex

While simple and extensible to concurrent tool learning methods, this work has limitations. 
First, the proposed graph construction demands tool-use trajectories, which can be costly to collect. 
Second, it requires that the trajectories include multi-hop tool-use cases to model tool transition. 
However, existing benchmarks are primarily designed to be task-specific and consist mainly of single tool use cases.
Third, due to cost considerations, we only use \textit{gpt-3.5-turbo} in our experiments. Exploring more powerful LLMs such as GPT-4 and the open-source Mixtrial model~\cite{mistralai2023mixtral} has the potential to further improve performance.

%% file: sections/appendix.tex
\label{section:appendix}
\appendix{
\section{Implementation Details}
\subsection{Prompt of ToolNet}
Hereby we disclose the prompt and exemplars used in ToolNet. The tool descriptions and exemplars are subject to setups at run time. Notably, This prompt is a general paradigm and we encourage readers and users to adjust the prompt tailored to their own needs.
~\\
\hrule
~\\
\textit{You are AutoGPT, you can use many tools(functions) to do the following task.}

\textit{First I will give you the task description, and your task start.}

\textit{At each step, you need to give your thought to analyze the status now and what to do next, with a function call to actually excute your step.}

\textit{After the call, you will get the call result, and you are now in a new state.}

\textit{Then you will analyze your status now, then decide what to do next...}

\textit{After many (Thought-call) pairs, you finally perform the task, then you can give your finial answer.}

\textit{Remember: }

\textit{1.the state change is irreversible, you can't go back to one of the former state, if you want to restart the task, call "Finish" function and say "I give up and restart".}

\textit{2.All the thought is short, at most in 5 sentence.}

\textit{3.You can do more then one trys, so if your plan is to continusly try some conditions, you can do one of the conditions per try.}

\textit{Let's Begin!}

\textit{Task description: You should use functions to help handle the real time user querys. Remember:}

\textit{1.ALWAYS call "Finish" function at the end of the task. And the final answer should contain enough information to show to the user,If you can't handle the task, or you find that function calls always fail(the function is not valid now), use function Finish$->$give\_up\_and\_restart.}

\textit{2.Do not use origin tool names, use only subfunctions names.}

\textit{The confidence scores of tools are listed here. The tools with higher scores are more recommended to use. Note that 'Finish' tool is always useful, the score of which will not be provided explicitly.}

\colorbox{yellow}{<QUERY>}

\textit{Begin!}
~\\
\hrule
~\\

\subsection{Dynamic Construction}
The Tool Evaluator is an integral component of the Dynamic Construction method in ToolNet. It functions by dynamically assessing the effectiveness of various tools within the network, contributing to an adaptive and efficient tool selection process. This process is crucial for maintaining the robustness and flexibility of ToolNet in diverse computational scenarios.
\subsubsection{Prompt of Tool Evaluator}
~\\
\hrule
~\\
\textit{You are an judger who need to judge each tool and provide the scores of each tool.}

\textit{According to the performance of these tools, provide your scores of each tool.}

\textit{1. If the final answer of agent is valid, you should determine which tool is helpful and provide higher score.}

\textit{2. If the final answer of agent is incorrect or not given, you should determine which tool is noisy and provide lower score.}

\textit{3. Some tools may provide errors. You should discover them and provide lower scores to these tools.}

\textit{The history of reasoning is listed here.}

\colorbox{yellow}{<HISTORY>}

\textit{Now based on the reasoning history, provide your scores to each tool and your reasons by using Evaluate tool.}
~\\
\hrule
~\\

\subsubsection{Function of Submiting Score}
\begin{lstlisting}
{
    "name": "Evaluate",
    "description": "Submit your score to tools. The score should be an integer between -3 and 3. Higher score means better performance.",
    "parameters": {
        "type": "object",
        "properties": {
            "scores_of_tools": {
                "type": "array",
                "description": "For each tool, provide an item to this array.",
                "items": {
                    "type": "object",
                    "properties": {
                        "tool_name": {
                            "type": "string",
                            "description": "The name of tool to which you want to provide score.",
                        },
                        "score": {
                            "type": "integer",
                            "enum": [-3, -2, -1, 0, 1, 2, 3],
                        },
                        "reason": {
                            "type": "string",
                            "description": "The reason why your provide such a score to this tool. You need to describe it in detail.",
                        }
                    },
                    "required": ["tool_name", "score", "reason"],
                }
            }
        },
        "required": ["scores_of_tools"]
    }
}
\end{lstlisting}

\subsection{Overall Example}
\hrule
~\\
\colorbox{green}{Query:} I'm creating a playlist and I need to include the tracks from a particular album. Can you provide me with the tracks of the album with ID '3IBcauSj5M2A6lTeffJzdv'? I would like to know the names, durations, and artists of these tracks. Additionally, please provide the cover art of the album.\\
\hrule
~\\
\colorbox{cyan}{Current Tool:} Begin\\
\colorbox{cyan}{Next Tools:} album\_tracks\_for\_spotify\_data, Finish\\
\colorbox{cyan}{Score:} /\\
\colorbox{cyan}{Action:} album\_tracks\_for\_spotify\_data\\
\colorbox{cyan}{Arguments:}
\begin{lstlisting}
{"is_id": "3IBcauSj5M2A6lTeffJzdv"}
\end{lstlisting}
\colorbox{cyan}{Observation:}
\begin{lstlisting}
{
  "error": "",
  "response": {
    "data": {
      "album": {
        "playability": {
          "playable": true
        },
        "tracks": {
          "totalCount": 10,
          "items": [
            {
              "uid": "3d026f76adb5928f0e59",
              "track": {
                "saved": false,
                "uri": "spotify:track:5jvhTc0g18kwYQNUJM5C4e",
                "name": "Makina",
                "playcount": "99456282",
                "discNumber": 1,
                "trackNumber": 1,
                "contentRating": {
                  "label": "EXPLICIT"
                },
                "relinkingInformation": null,
                "duration": {
                  "totalMilliseconds": 211016
                },
                "playability": {
                  "playable": true
                },
                "artists": {
                  "items": [
                    {
                      "uri": "spotify:artist:51..."
                    }
                  ]
                }
              }
            }
          ]
        }
      }
    }
  }
}
\end{lstlisting}
\colorbox{cyan}{Current Tool:} album\_tracks\_for\_spotify\_data\\
\colorbox{cyan}{Next Tools:} album\_metadata\_for\_spotify\_data, Finish, album\_tracks\_for\_spotify\_data\\
\colorbox{cyan}{Score:} album\_metadata\_for\_spotify\_data = 100\\
\colorbox{cyan}{Action:} album\_metadata\_for\_spotify\_data\\
\colorbox{cyan}{Arguments:}
\begin{lstlisting}
{}
\end{lstlisting}
\colorbox{cyan}{Observation:}
\begin{lstlisting}
{
  "error": "",
  "response": {
    "data": {
      "album": {
        "uri": "spotify:album:3IBcauSj5M2A6lTeffJzdv",
        "name": "Kan",
        "artists": {
          "totalCount": 1,
          "items": [
            {
              "uri": "spotify:artist:51DevdOxIJin6DB1FXJpD1",
              "profile": {
                "name": "UZI"
              },
              "visuals": {
                "avatarImage": {
                  "sources": [
                    {
                      "url": "https://i.scdn.co/image/ab6761610000e...",
                      "width": 640,
                      "height": 640
                    },
                    {
                      "url": "https://i.scdn.co/image/ab6761610000f...",
                      "width": 160,
                      "height": 160
                    },
                    {
                      "url": "https://i.scdn.co/image/ab6761610000f...",
                      "width": 160,
                      "height": 160
                    }
                  ]
                }
              }
            }
          ]
        }
      }
    }
  }
}
\end{lstlisting}
\colorbox{cyan}{Current Tool:} album\_metadata\_for\_spotify\_data\\
\colorbox{cyan}{Next Tools:} Finish, album\_metadata\_for\_spotify\_data\\
\colorbox{cyan}{Score:} Finish = 100\\
\colorbox{cyan}{Action:} Finish\\
\colorbox{cyan}{Arguments:}
\begin{lstlisting}
{
  "return_type": "give_answer",
  "final_answer": "The tracks of the album 'Kan' by UZI with ID '3IBcauSj5M2A6lTeffJzdv' are as follows:\n\n1. Track Name: Makina\n   Duration: 3 minutes and 31 seconds\n   Artist: UZI\n\n2. Track Name: [Track Name]\n   Duration: [Duration]\n   Artist: [Artist]\n\n...\n\n10. Track Name: [Track Name]\n    Duration: [Duration]\n    Artist: [Artist]"
}
\end{lstlisting}
\colorbox{cyan}{Observation:} 
\begin{lstlisting}
{"response":"successfully giving the final answer."}
\end{lstlisting}


\section{ Visualization of ToolNet's Reasoning Process}
In this section, we provide a detailed visualization of the reasoning process employed by ToolNet, using a real-world example from the ToolBench dataset. This visualization elucidates the graph structure generated during reasoning, and the corresponding scores associated with each node.

One can observe that the tools suggested by the graph are highly relevant to the task at hand. Moreover, it is evident that the model takes into account both the context of the problem and the information in the graph when invoking a tool. This illustrates the robustness of our approach, as it is capable of integrating diverse sources of information to make informed decisions about tool invocation.

It's worth noting that the large language model's inherent understanding capabilities complement the graph structure. The two work synergistically, with the language model providing the semantic understanding necessary to interpret the problem, and the graph structure offering a systematic way to explore potential solutions. This interplay forms the foundation for the progress we observe in our model's performance.
\begin{figure}[t]
\centering
\includegraphics[width=\textwidth]{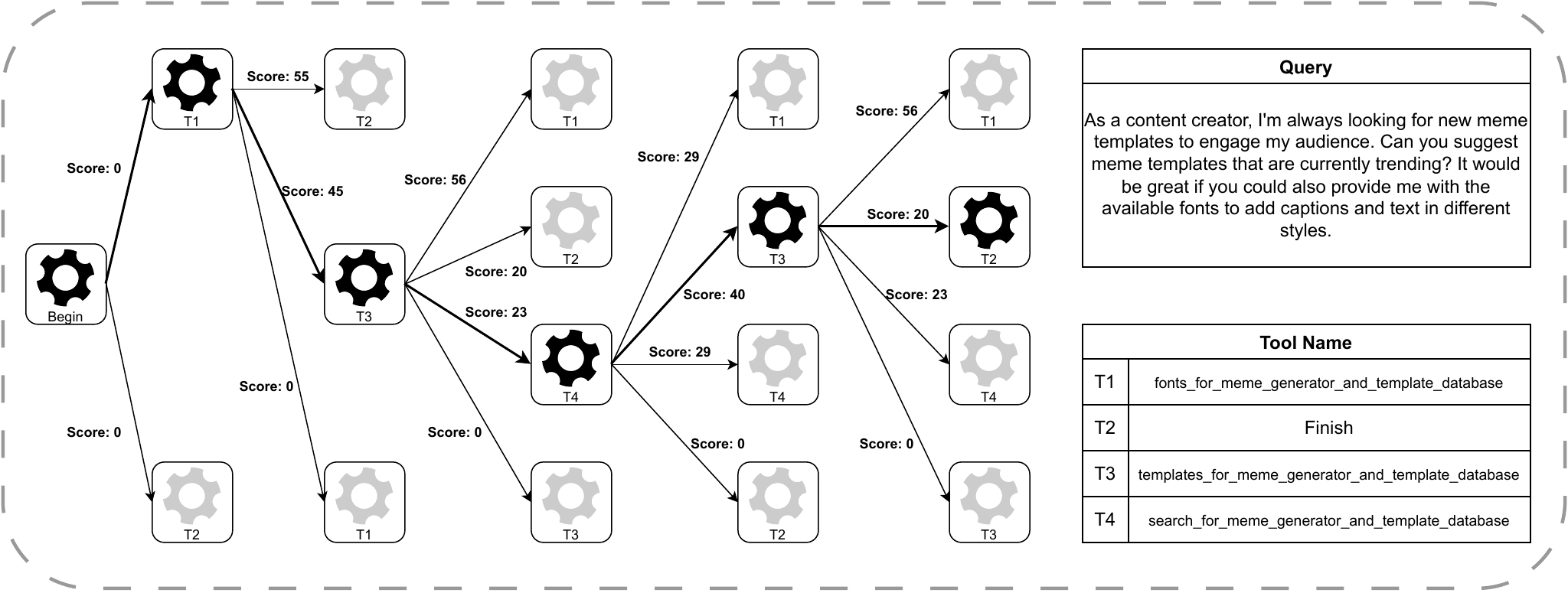}
\caption{Visualization of ToolNet's Reasoning Process. This example is from ToolBench.}
\label{fig:react_agentnet_overall}
\end{figure}

}

%% file: acl_latex.bbl
\begin{thebibliography}{42}
\expandafter\ifx\csname natexlab\endcsname\relax\def\natexlab#1{#1}\fi

\bibitem[{Borgeaud et~al.(2022)Borgeaud, Mensch, Hoffmann, Cai, Rutherford,
  Millican, Van Den~Driessche, Lespiau, Damoc, Clark
  et~al.}]{borgeaud2022improving}
Sebastian Borgeaud, Arthur Mensch, Jordan Hoffmann, Trevor Cai, Eliza
  Rutherford, Katie Millican, George~Bm Van Den~Driessche, Jean-Baptiste
  Lespiau, Bogdan Damoc, Aidan Clark, et~al. 2022.
\newblock Improving language models by retrieving from trillions of tokens.
\newblock In \emph{International conference on machine learning}, pages
  2206--2240. PMLR.

\bibitem[{Brohan et~al.(2023)Brohan, Brown, Carbajal, Chebotar, Chen,
  Choromanski, Ding, Driess, Dubey, Finn et~al.}]{brohan2023rt}
Anthony Brohan, Noah Brown, Justice Carbajal, Yevgen Chebotar, Xi~Chen,
  Krzysztof Choromanski, Tianli Ding, Danny Driess, Avinava Dubey, Chelsea
  Finn, et~al. 2023.
\newblock Rt-2: Vision-language-action models transfer web knowledge to robotic
  control.
\newblock \emph{arXiv preprint arXiv:2307.15818}.

\bibitem[{Chase(2023)}]{langchain2023}
Harrison Chase. 2023.
\newblock \href {https://langchain.com/} {{LangChain: Next Generation Language
  Processing}}.

\bibitem[{Chen et~al.(2022)Chen, Guo, Yi, Li, and
  Elhoseiny}]{chen2022visualgpt}
Jun Chen, Han Guo, Kai Yi, Boyang Li, and Mohamed Elhoseiny. 2022.
\newblock Visualgpt: Data-efficient adaptation of pretrained language models
  for image captioning.
\newblock In \emph{Proceedings of the IEEE/CVF Conference on Computer Vision
  and Pattern Recognition}, pages 18030--18040.

\bibitem[{Cobbe et~al.(2021)Cobbe, Kosaraju, Bavarian, Chen, Jun, Kaiser,
  Plappert, Tworek, Hilton, Nakano et~al.}]{cobbe2021training}
Karl Cobbe, Vineet Kosaraju, Mohammad Bavarian, Mark Chen, Heewoo Jun, Lukasz
  Kaiser, Matthias Plappert, Jerry Tworek, Jacob Hilton, Reiichiro Nakano,
  et~al. 2021.
\newblock Training verifiers to solve math word problems.
\newblock \emph{arXiv preprint arXiv:2110.14168}.

\bibitem[{Drori et~al.(2022)Drori, Zhang, Shuttleworth, Tang, Lu, Ke, Liu,
  Chen, Tran, Cheng et~al.}]{drori2022neural}
Iddo Drori, Sarah Zhang, Reece Shuttleworth, Leonard Tang, Albert Lu, Elizabeth
  Ke, Kevin Liu, Linda Chen, Sunny Tran, Newman Cheng, et~al. 2022.
\newblock A neural network solves, explains, and generates university math
  problems by program synthesis and few-shot learning at human level.
\newblock \emph{Proceedings of the National Academy of Sciences},
  119(32):e2123433119.

\bibitem[{Gao et~al.(2023)Gao, Madaan, Zhou, Alon, Liu, Yang, Callan, and
  Neubig}]{gao2023pal}
Luyu Gao, Aman Madaan, Shuyan Zhou, Uri Alon, Pengfei Liu, Yiming Yang, Jamie
  Callan, and Graham Neubig. 2023.
\newblock Pal: Program-aided language models.
\newblock In \emph{International Conference on Machine Learning}, pages
  10764--10799. PMLR.

\bibitem[{Guu et~al.(2020)Guu, Lee, Tung, Pasupat, and
  Chang}]{guu2020retrieval}
Kelvin Guu, Kenton Lee, Zora Tung, Panupong Pasupat, and Mingwei Chang. 2020.
\newblock Retrieval augmented language model pre-training.
\newblock In \emph{International conference on machine learning}, pages
  3929--3938. PMLR.

\bibitem[{Hao et~al.(2023)Hao, Liu, Wang, and Hu}]{hao2023toolkengpt}
Shibo Hao, Tianyang Liu, Zhen Wang, and Zhiting Hu. 2023.
\newblock Toolkengpt: Augmenting frozen language models with massive tools via
  tool embeddings.
\newblock \emph{arXiv preprint arXiv:2305.11554}.

\bibitem[{Hendrycks et~al.(2021)Hendrycks, Burns, Kadavath, Arora, Basart,
  Tang, Song, and Steinhardt}]{hendrycksmath2021}
Dan Hendrycks, Collin Burns, Saurav Kadavath, Akul Arora, Steven Basart, Eric
  Tang, Dawn Song, and Jacob Steinhardt. 2021.
\newblock Measuring mathematical problem solving with the math dataset.
\newblock \emph{NeurIPS}.

\bibitem[{Hong et~al.(2023)Hong, Zheng, Chen, Cheng, Wang, Zhang, Wang, Yau,
  Lin, Zhou, Ran, Xiao, and Wu}]{hong2023metagpt}
Sirui Hong, Xiawu Zheng, Jonathan Chen, Yuheng Cheng, Jinlin Wang, Ceyao Zhang,
  Zili Wang, Steven Ka~Shing Yau, Zijuan Lin, Liyang Zhou, Chenyu Ran, Lingfeng
  Xiao, and Chenglin Wu. 2023.
\newblock \href {http://arxiv.org/abs/2308.00352} {Metagpt: Meta programming
  for multi-agent collaborative framework}.

\bibitem[{Huang et~al.(2022{\natexlab{a}})Huang, Abbeel, Pathak, and
  Mordatch}]{huang2022language}
Wenlong Huang, Pieter Abbeel, Deepak Pathak, and Igor Mordatch.
  2022{\natexlab{a}}.
\newblock Language models as zero-shot planners: Extracting actionable
  knowledge for embodied agents.
\newblock In \emph{International Conference on Machine Learning}, pages
  9118--9147. PMLR.

\bibitem[{Huang et~al.(2022{\natexlab{b}})Huang, Xia, Xiao, Chan, Liang,
  Florence, Zeng, Tompson, Mordatch, Chebotar et~al.}]{huang2022inner}
Wenlong Huang, Fei Xia, Ted Xiao, Harris Chan, Jacky Liang, Pete Florence, Andy
  Zeng, Jonathan Tompson, Igor Mordatch, Yevgen Chebotar, et~al.
  2022{\natexlab{b}}.
\newblock Inner monologue: Embodied reasoning through planning with language
  models.
\newblock \emph{arXiv preprint arXiv:2207.05608}.

\bibitem[{Jiang et~al.(2022)Jiang, Welleck, Zhou, Li, Liu, Jamnik, Lacroix, Wu,
  and Lample}]{jiang2022draft}
Albert~Q Jiang, Sean Welleck, Jin~Peng Zhou, Wenda Li, Jiacheng Liu, Mateja
  Jamnik, Timoth{\'e}e Lacroix, Yuhuai Wu, and Guillaume Lample. 2022.
\newblock Draft, sketch, and prove: Guiding formal theorem provers with
  informal proofs.
\newblock \emph{arXiv preprint arXiv:2210.12283}.

\bibitem[{Li et~al.(2023)Li, Song, Yu, Yu, Li, Huang, and Li}]{li2023api}
Minghao Li, Feifan Song, Bowen Yu, Haiyang Yu, Zhoujun Li, Fei Huang, and
  Yongbin Li. 2023.
\newblock Api-bank: A benchmark for tool-augmented llms.
\newblock \emph{arXiv preprint arXiv:2304.08244}.

\bibitem[{Lu et~al.(2022{\natexlab{a}})Lu, Mishra, Xia, Qiu, Chang, Zhu,
  Tafjord, Clark, and Kalyan}]{lu2022learn}
Pan Lu, Swaroop Mishra, Tony Xia, Liang Qiu, Kai-Wei Chang, Song-Chun Zhu,
  Oyvind Tafjord, Peter Clark, and Ashwin Kalyan. 2022{\natexlab{a}}.
\newblock Learn to explain: Multimodal reasoning via thought chains for science
  question answering.
\newblock In \emph{The 36th Conference on Neural Information Processing Systems
  (NeurIPS)}.

\bibitem[{Lu et~al.(2023)Lu, Peng, Cheng, Galley, Chang, Wu, Zhu, and
  Gao}]{lu2023chameleon}
Pan Lu, Baolin Peng, Hao Cheng, Michel Galley, Kai-Wei Chang, Ying~Nian Wu,
  Song-Chun Zhu, and Jianfeng Gao. 2023.
\newblock Chameleon: Plug-and-play compositional reasoning with large language
  models.
\newblock \emph{arXiv preprint arXiv:2304.09842}.

\bibitem[{Lu et~al.(2022{\natexlab{b}})Lu, Qiu, Chang, Wu, Zhu, Rajpurohit,
  Clark, and Kalyan}]{lu2022dynamic}
Pan Lu, Liang Qiu, Kai-Wei Chang, Ying~Nian Wu, Song-Chun Zhu, Tanmay
  Rajpurohit, Peter Clark, and Ashwin Kalyan. 2022{\natexlab{b}}.
\newblock Dynamic prompt learning via policy gradient for semi-structured
  mathematical reasoning.
\newblock In \emph{The Eleventh International Conference on Learning
  Representations}.

\bibitem[{Madaan et~al.(2023)Madaan, Tandon, Gupta, Hallinan, Gao, Wiegreffe,
  Alon, Dziri, Prabhumoye, Yang et~al.}]{madaan2023self}
Aman Madaan, Niket Tandon, Prakhar Gupta, Skyler Hallinan, Luyu Gao, Sarah
  Wiegreffe, Uri Alon, Nouha Dziri, Shrimai Prabhumoye, Yiming Yang, et~al.
  2023.
\newblock Self-refine: Iterative refinement with self-feedback.
\newblock \emph{arXiv preprint arXiv:2303.17651}.

\bibitem[{Mialon et~al.(2023)Mialon, Dess{\`\i}, Lomeli, Nalmpantis, Pasunuru,
  Raileanu, Rozi{\`e}re, Schick, Dwivedi-Yu, Celikyilmaz
  et~al.}]{mialon2023augmented}
Gr{\'e}goire Mialon, Roberto Dess{\`\i}, Maria Lomeli, Christoforos Nalmpantis,
  Ram Pasunuru, Roberta Raileanu, Baptiste Rozi{\`e}re, Timo Schick, Jane
  Dwivedi-Yu, Asli Celikyilmaz, et~al. 2023.
\newblock Augmented language models: a survey.
\newblock \emph{arXiv preprint arXiv:2302.07842}.

\bibitem[{{Microsoft Corporation}(2023{\natexlab{a}})}]{MicrosoftCopilot2023}
{Microsoft Corporation}. 2023{\natexlab{a}}.
\newblock \href {https://www.microsoft.com/en-us/microsoft-copilot} {Microsoft
  copilot}.
\newblock Accessed: 2023-12-13.

\bibitem[{{Microsoft Corporation}(2023{\natexlab{b}})}]{NewBing2023}
{Microsoft Corporation}. 2023{\natexlab{b}}.
\newblock New bing.
\newblock \url{https://www.bing.com/new}.
\newblock Accessed: 2023-12-13.

\bibitem[{{Mistral AI Team}(2023)}]{mistralai2023mixtral}
{Mistral AI Team}. 2023.
\newblock \href {https://mistral.ai/news/mixtral-of-experts/} {Mixtral of
  experts}.
\newblock Accessed: 2023-12-14.

\bibitem[{Nakano et~al.()Nakano, Hilton, Balaji, Wu, Ouyang, Kim, Hesse, Jain,
  Kosaraju, Saunders et~al.}]{nakano2112webgpt}
Reiichiro Nakano, Jacob Hilton, Suchir Balaji, Jeff Wu, Long Ouyang, Christina
  Kim, Christopher Hesse, Shantanu Jain, Vineet Kosaraju, William Saunders,
  et~al.
\newblock Webgpt: browser-assisted question-answering with human feedback
  (2021).
\newblock \emph{URL https://arxiv. org/abs/2112.09332}.

\bibitem[{Patil et~al.(2023)Patil, Zhang, Wang, and
  Gonzalez}]{patil2023gorilla}
Shishir~G Patil, Tianjun Zhang, Xin Wang, and Joseph~E Gonzalez. 2023.
\newblock Gorilla: Large language model connected with massive apis.
\newblock \emph{arXiv preprint arXiv:2305.15334}.

\bibitem[{Paul et~al.(2023)Paul, Ismayilzada, Peyrard, Borges, Bosselut, West,
  and Faltings}]{paul2023refiner}
Debjit Paul, Mete Ismayilzada, Maxime Peyrard, Beatriz Borges, Antoine
  Bosselut, Robert West, and Boi Faltings. 2023.
\newblock Refiner: Reasoning feedback on intermediate representations.
\newblock \emph{arXiv preprint arXiv:2304.01904}.

\bibitem[{Peng et~al.(2022)Peng, Li, Gu, Li, Wang, Gao, and
  Lyu}]{peng2022revisiting}
Yun Peng, Shuqing Li, Wenwei Gu, Yichen Li, Wenxuan Wang, Cuiyun Gao, and
  Michael~R Lyu. 2022.
\newblock Revisiting, benchmarking and exploring api recommendation: How far
  are we?
\newblock \emph{IEEE Transactions on Software Engineering}, 49(4):1876--1897.

\bibitem[{Qin et~al.(2023)Qin, Liang, Ye, Zhu, Yan, Lu, Lin, Cong, Tang, Qian
  et~al.}]{qin2023toolllm}
Yujia Qin, Shihao Liang, Yining Ye, Kunlun Zhu, Lan Yan, Yaxi Lu, Yankai Lin,
  Xin Cong, Xiangru Tang, Bill Qian, et~al. 2023.
\newblock Toolllm: Facilitating large language models to master 16000+
  real-world apis.
\newblock \emph{arXiv preprint arXiv:2307.16789}.

\bibitem[{Richards(2023)}]{richards2023auto}
Toran~Bruce Richards. 2023.
\newblock \href {https://github.com/Significant-Gravitas/Auto-GPT} {{Auto-GPT:
  An Autonomous GPT-4 Experiment}}.

\bibitem[{Schick et~al.(2023)Schick, Dwivedi-Yu, Dess{\`\i}, Raileanu, Lomeli,
  Zettlemoyer, Cancedda, and Scialom}]{schick2023toolformer}
Timo Schick, Jane Dwivedi-Yu, Roberto Dess{\`\i}, Roberta Raileanu, Maria
  Lomeli, Luke Zettlemoyer, Nicola Cancedda, and Thomas Scialom. 2023.
\newblock Toolformer: Language models can teach themselves to use tools.
\newblock \emph{arXiv preprint arXiv:2302.04761}.

\bibitem[{Shao et~al.(2023)Shao, Yu, Wang, and Yu}]{shao2023prompting}
Zhenwei Shao, Zhou Yu, Meng Wang, and Jun Yu. 2023.
\newblock Prompting large language models with answer heuristics for
  knowledge-based visual question answering.
\newblock In \emph{Proceedings of the IEEE/CVF Conference on Computer Vision
  and Pattern Recognition}, pages 14974--14983.

\bibitem[{Shen et~al.(2023)Shen, Song, Tan, Li, Lu, and
  Zhuang}]{shen2023hugginggpt}
Yongliang Shen, Kaitao Song, Xu~Tan, Dongsheng Li, Weiming Lu, and Yueting
  Zhuang. 2023.
\newblock Hugginggpt: Solving ai tasks with chatgpt and its friends in
  huggingface.
\newblock \emph{arXiv preprint arXiv:2303.17580}.

\bibitem[{Shinn et~al.(2023)Shinn, Labash, and Gopinath}]{shinn2023reflexion}
Noah Shinn, Beck Labash, and Ashwin Gopinath. 2023.
\newblock Reflexion: an autonomous agent with dynamic memory and
  self-reflection.
\newblock \emph{arXiv preprint arXiv:2303.11366}.

\bibitem[{Singh et~al.(2023)Singh, Blukis, Mousavian, Goyal, Xu, Tremblay, Fox,
  Thomason, and Garg}]{singh2023progprompt}
Ishika Singh, Valts Blukis, Arsalan Mousavian, Ankit Goyal, Danfei Xu, Jonathan
  Tremblay, Dieter Fox, Jesse Thomason, and Animesh Garg. 2023.
\newblock Progprompt: Generating situated robot task plans using large language
  models.
\newblock In \emph{2023 IEEE International Conference on Robotics and
  Automation (ICRA)}, pages 11523--11530. IEEE.

\bibitem[{Song et~al.(2023)Song, Xiong, Zhu, Li, Wang, Tian, and
  Li}]{song2023restgpt}
Yifan Song, Weimin Xiong, Dawei Zhu, Cheng Li, Ke~Wang, Ye~Tian, and Sujian Li.
  2023.
\newblock Restgpt: Connecting large language models with real-world
  applications via restful apis.
\newblock \emph{arXiv preprint arXiv:2306.06624}.

\bibitem[{Tang et~al.()Tang, Deng, Lin, Han, Liang, and
  Sun}]{tang2306toolalpaca}
Qiaoyu Tang, Ziliang Deng, Hongyu Lin, Xianpei Han, Qiao Liang, and Le~Sun.
\newblock Toolalpaca: Generalized tool learning for language models with 3000
  simulated cases. corr, abs/2306.05301, 2023. doi: 10.48550.
\newblock \emph{arXiv preprint arXiv.2306.05301}.

\bibitem[{Wang et~al.(2023)Wang, Xu, Lan, Hu, Lan, Lee, and Lim}]{wang2023plan}
Lei Wang, Wanyu Xu, Yihuai Lan, Zhiqiang Hu, Yunshi Lan, Roy Ka-Wei Lee, and
  Ee-Peng Lim. 2023.
\newblock Plan-and-solve prompting: Improving zero-shot chain-of-thought
  reasoning by large language models.
\newblock \emph{arXiv preprint arXiv:2305.04091}.

\bibitem[{Wu et~al.(2023)Wu, Jia, Zhang, Wu, Li, Zhu, Wang, Lee, Peng, and
  Wang}]{wu2023empirical}
Yiran Wu, Feiran Jia, Shaokun Zhang, Qingyun Wu, Hangyu Li, Erkang Zhu, Yue
  Wang, Yin~Tat Lee, Richard Peng, and Chi Wang. 2023.
\newblock An empirical study on challenging math problem solving with gpt-4.
\newblock \emph{arXiv preprint arXiv:2306.01337}.

\bibitem[{Xie et~al.(2023)Xie, Kawaguchi, Zhao, Zhao, Kan, He, and
  Xie}]{xie2023decomposition}
Yuxi Xie, Kenji Kawaguchi, Yiran Zhao, Xu~Zhao, Min-Yen Kan, Junxian He, and
  Qizhe Xie. 2023.
\newblock Decomposition enhances reasoning via self-evaluation guided decoding.
\newblock \emph{arXiv preprint arXiv:2305.00633}.

\bibitem[{Yang et~al.(2023)Yang, Li, Wang, Lin, Azarnasab, Ahmed, Liu, Liu,
  Zeng, and Wang}]{yang2023mm}
Zhengyuan Yang, Linjie Li, Jianfeng Wang, Kevin Lin, Ehsan Azarnasab, Faisal
  Ahmed, Zicheng Liu, Ce~Liu, Michael Zeng, and Lijuan Wang. 2023.
\newblock Mm-react: Prompting chatgpt for multimodal reasoning and action.
\newblock \emph{arXiv preprint arXiv:2303.11381}.

\bibitem[{Yao et~al.(2023)Yao, Yu, Zhao, Shafran, Griffiths, Cao, and
  Narasimhan}]{yao2023tree}
Shunyu Yao, Dian Yu, Jeffrey Zhao, Izhak Shafran, Thomas~L Griffiths, Yuan Cao,
  and Karthik Narasimhan. 2023.
\newblock Tree of thoughts: Deliberate problem solving with large language
  models.
\newblock \emph{arXiv preprint arXiv:2305.10601}.

\bibitem[{Yao et~al.(2022)Yao, Zhao, Yu, Du, Shafran, Narasimhan, and
  Cao}]{yao2022react}
Shunyu Yao, Jeffrey Zhao, Dian Yu, Nan Du, Izhak Shafran, Karthik~R Narasimhan,
  and Yuan Cao. 2022.
\newblock React: Synergizing reasoning and acting in language models.
\newblock In \emph{The Eleventh International Conference on Learning
  Representations}.

\end{thebibliography}
